\documentclass[runningheads]{llncs}
\usepackage{graphicx}
\usepackage{wrapfig}
\usepackage[usenames,dvipsnames]{color}

\begin{document}
\title{Safe Reinforcement Learning in a Simulated Robotic Arm}
\author{Luka Kova\v{c}\inst{1}
\and
Igor Farka\v{s}\inst{2}}
\authorrunning{L. Kova\v{c} and I. Farka\v{s}}
\institute{Faculty of Computer and Information Science, University of Ljubljana, Slovenia \and
Department of Applied Informatics, Comenius University Bratislava, Slovakia
\email{lk1114@student.uni-lj.si, igor.farkas@fmph.uniba.sk}}
\maketitle              
\begin{abstract}
Reinforcement learning (RL) agents need to explore their environments in order to learn optimal policies.
In many environments and tasks, safety is of critical importance. The widespread use of simulators offers a number of advantages, including safe exploration which will be inevitable in cases when RL systems need to be trained directly in the physical environment (e.g. in human-robot interaction). The popular Safety Gym library offers three mobile agent types that can learn goal-directed tasks while considering various safety constraints.
In this paper, we extend the applicability of safe RL algorithms by creating a customized environment with Panda robotic arm where Safety Gym algorithms can be tested. We performed pilot experiments with the popular PPO algorithm comparing the baseline with the constrained version and show that the constrained version is able to learn the equally good policy while better complying with safety constraints and taking longer training time as expected. 

\end{abstract}
\section{Introduction}

Reinforcement learning (RL) L agents need to explore their environments to learn optimal behaviours. Sometimes an agent might perform a dangerous action, therefore exploration is risky. Safe RL can be defined as the process of learning to maximize the reward and at the same time to ensure respecting safety constraints during learning \cite{garcia2015}. It is usually possible to train the agent in a simulated environment, and then after learning to transfer the learned policy to a physical agent in the real world. However, because of difficulties in simulating certain behaviours (e.g. human interaction, real-world scenarios in traffic, etc.) agent’s learning is transferred to the real world, where safety concerns are of great importance. 

To address these problems, OpenAI created Safety Gym, a suite of environments and tools for measuring progress toward RL agents that respect safety constraints while learning \cite{ray2019}, not only in testing. Safety Gym offers three different agent types (point, car, quadruped), different tasks (goal, button, push) and different safety constraints (hazards, vases, etc.). With those tools, one can create different layouts for trying out novel RL algorithms and having a common ground for benchmarking and evaluating them.

Our work integrates a new model of an agent (a robotic arm) into the Safety Gym environment. In a simulated environment, we are able to evaluate the agent’s behaviour regarding the safety concerns. Research in this direction can produce significant contributions into human-robot interaction in the future. 

An optimal policy in constrained RL is given by:
\begin{equation}
\pi^{*} = \arg \max_{\pi \in {\rm \Pi}_C} J_r(\pi)
\quad \quad \quad {\rm \Pi}_C = \{ \pi : J_{c_i}(\pi) \le d_i, \ i = 1,...,k\}
\label{eq1}
\end{equation}
where $J_r(\pi)$ is a reward-based objective function and each $J_{c_i}$ is a cost-based constraint function, involving thresholds $d_i$ (a human-selected hyperparameters). These constraint functions form a feasible set (of allowable policies) ${\rm \Pi}_C$ that has been defined in the framework of constrained Markov Decision Processes \cite{altman99}.
In our case, $d_i = 1$, if the arm collides with the obstacle, otherwise it is 0. Hence, Lagrangian method uses a two-component loss function (reward-based and cost-based). In eq.~\ref{eq1}, the cost-based component is included within the space of acceptable policies ${\rm \Pi}_C$. The optimization problem can also be expressed  as
$$
\max_{\theta}\min_{\lambda\ge 0} L(\phi,\theta) \doteq f(\theta) - \lambda g(\theta)
$$
where the two terms of the loss function correspond to the reward and the cost, involving policy network parameters $\theta$ and Lagrangian hyperparameter $\lambda$ \cite{ray2019}.

\section{Finding a technical solution}

On one hand, it is positive that there exist various Python libraries and robotic simulators built on a variety of physics simulation engines. On the other hand, combining them or making extensions may often not be easy. 
Our primary motivation was to integrate safe RL algorithms with a robotic arm (not included in the Safety Gym library) that can be used in human-robot interaction. Finding a solution was not straightforward, though. The integration could be achieved in two ways: (1) Bringing a robotic arm model into Safety Gym framework, or (2) using a different or a customized environment with a robotic arm and integrate just the safety algorithms into it. This led us to the exploration of feasible options. 

Safety Gym is built on the MuJoCo physics engine \cite{Todorov12}, so we first tried to import a Reacher model (a simplified robotic arm) from OpenAI Gym to Safety Gym. This should be compatible, since both are based on MuJoCo. But various technical problems 
(a lot of dependencies, the need to use older versions of Python and Tensorflow) discouraged us from pursuing this line of investigation. 

Within the second option, we tried to connect Safety Gym with commonly used robotic simulator CoppeliaSim using PyRep library built on top of it -- but this did not work due to incompatibility issues.

Finally, we used a PyBullet physics simulation engine that is built with python and is an open source project, so it is well documented and with a lot of examples already on the web. That helped a lot to set up the environment in the desirable way. 
Because there are already a lot of examples, we found the environment with a robotic arm that is implemented with PyBullet and is compatible with OpenAI Gym - panda-gym. 
Our source code with the installation guide and instructions of how to run the environment can be found here.\footnote{\url{https://github.com/lukakovac99/robotic-arm-safeRL}}
We also implemented two aditional arms to the environment -- xarm and kuka -- that can be used for training with safety algorithms.

\section{Experiments}

We used the Proximal Policy Optimization (PPO), a well-known efficient policy gradient method for RL \cite{schulman2017} in our pilot experiments.
We compared the basic PPO with its constrained version (cPPO) using the panda-gym robotics arm (with 7 DoF).

\begin{wrapfigure}{r}{0.35\textwidth}
\vspace{-8mm}
\includegraphics[width=0.33\textwidth]{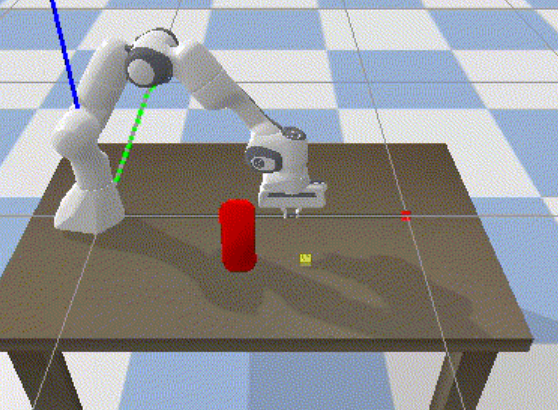}
\caption{Panda arm learned to reach the target (yellow cube) without colliding with an obstacle (red) in front of it.} 
\label{fig1}
\vspace{-7mm}
\end{wrapfigure}

Regarding the action representation, we considered two options: (a) in PyBullet, the action representation is given by a vector [$dx, dy, dz$] which means changes of the tip of the arm in 3D Cartesian space (we label it AR1). Those values are used to calculate the new position of the tip and via inverse kinematics to calculate how much the joints should change.
(b) We also tested a ``classical'' actor output representation computed directly in the joint space as a 7-dim. vector of DoF angle changes in each step (AR2). These values are then directly added to move the arm (forward kinematics). We used dense reward hence simulating robotic vision enabling the robor to estimate the distance between the tip and the target, which served as information for calculating the (inversely proportional) reward.
Last but not least, we added an obstacle on the table in front of the target object (see Fig.\ref{fig1}).

\begin{wraptable}{r}{0.35\textwidth}
\vspace{-12mm}
\caption{Average cost (with std) per one run of the classical PPO algorithm and its constrained version in case of Panda arm reaching for a target, using two action representation formats.}
\label{tab1}
\begin{tabular}{lll}
 & PPO & cPPO \\
\hline
3D \, & 17.6$\pm$1.3 \, & \, 11.9$\pm$3.6 \\
7DoF \, & 23.8$\pm$5.0 \, & \, 17.0$\pm$1.9 \\
\hline
\end{tabular}
\vspace{-5mm}
\end{wraptable}

\begin{figure}
\includegraphics[width=\textwidth]{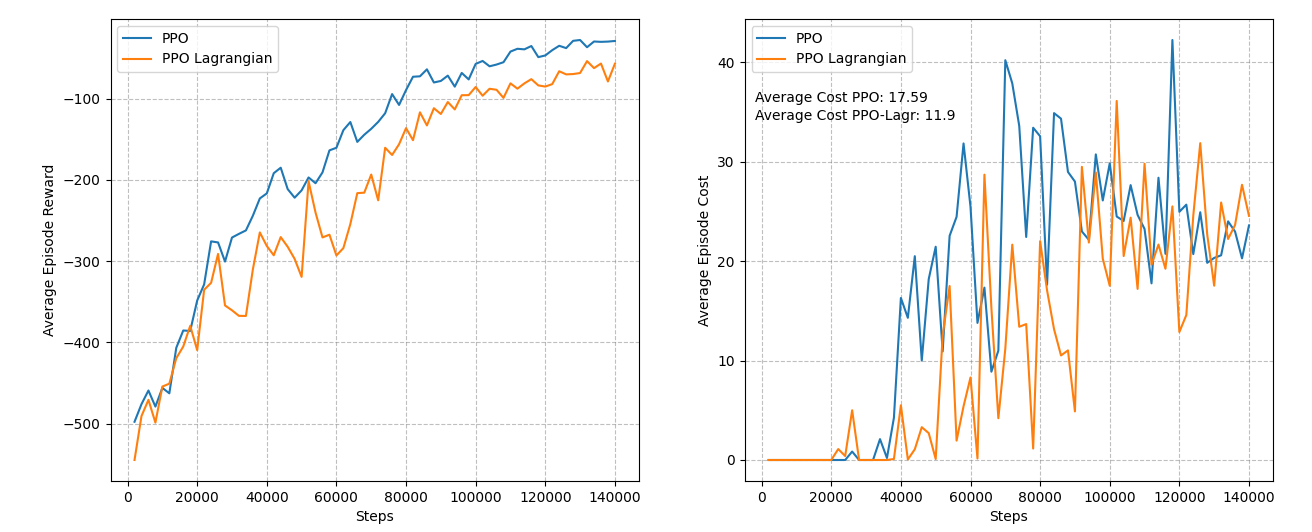}
\caption{Comparison of PPO and cPPO using panda arm in terms of reward (left) and cost (right). Constrained PPO is slower in learning and reaching the reward. On the other hand, it is keeping the cost at lower values hence making the arm behavior safer.} 
\label{fig2}
\end{figure}

In our four experiments (AR1/2, c/PPO) we used separate feedforward MLP policy networks with two hidden layers, each with 64 neurons, 1000 steps per epoch, maximum 200 epochs of training, and maximum number of steps per episode = 500.
The experiments lead to two observations (see Table~\ref{tab1}): (a) Regarding AR type, the agent learns faster (roughly with speedup factor of 2) and easier when using AR1 than AR2 (this is probably due to higher dimensionality of the state vector in the latter case). 
(b) Regarding the algorithm, cPPO yield lower average costs for both AR types. This makes Lagrangian PPO safer, with a tradeoff for length of training. Performance of both algorithms in case of AR1 is illustrated in Fig.~\ref{fig2}.

\section{Conclusion}

We presented pilot results with a robotic arm (panda gym) environment that is compatible with OpenAI Safety Gym, and verified the correct functionality on a selected algorithm (PPO). Constrained (Lagrangian) PPO algorithm was observed to have a longer learning time, but eventually learned the policies at the same level of efficiency while being all the way safer. 

The available code provides opportunities for experimenting with the robotic arm in various setups, trying also other algorithms available in Safety Gym  (TRPO, cTRPO and CPO), adding a proper obstacle representation, obstacle generation methods, or developing different safe tasks for the agent to perform.
\\

\noindent {\bf Acknowledgement:}
L.K. was supported by Erasmus mobility stipend, and I.F. by the Horizon Europe project TERAIS, no. 101079338 and by the national project APVV-21-0105.

\bibliographystyle{splncs04}
\bibliography{paper}
\end{document}